\documentclass[extended-abstract]{ccn}  
\usepackage{times}
\usepackage{soul}
\usepackage{url}
\usepackage[small]{caption}
\usepackage{graphicx}
\usepackage{amsmath}
\usepackage{amsthm}
\usepackage{booktabs}
\usepackage{algorithm}
\usepackage{algorithmic}

\title{A Minimal Model of Bounded Trade-Off Screening in Multi-Attribute Choice}

\author{%
  Manisha Dubey\affmark{1} \And
  Anirban Sarkar\affmark{2} \And
  Subramanian Ramamoorthy\affmark{1} \And
}
\affiliation{1}{School of Informatics, University of Edinburgh, UK}
\affiliation{2}{Cold Spring Harbor Laboratory, USA}
\emails{mdubey@ed.ac.uk, asarkar@cshl.edu, s.ramamoorthy@ed.ac.uk}

\begin{document}

\maketitle

\begin{abstract}
Human decision-making often involves choosing between multi-attribute alternatives, yet classical models assume fully compensatory utility aggregation despite evidence that people reject options with poor performance on critical attributes. We propose a bounded trade-off reasoning framework in which decisions are governed by a screening process that evaluates the balance between gains and losses across attributes. The model introduces a trade-off tolerance parameter that controls acceptable imbalance and can vary across contexts. Through simulation, we show that this mechanism produces preference patterns that differ from standard utility-based models and captures context-dependent variation in trade-off behavior. These results establish bounded trade-off screening as a plausible computational mechanism for multi-attribute choice and generate testable predictions for future behavioral studies.
\end{abstract}

\section{Introduction}
Many real-world decisions involve choosing between alternatives that differ along multiple attributes such as cost, time, risk, and quality. Classical decision theory assumes that attributes can be aggregated into a single scalar utility, allowing gains in one dimension to compensate for losses in another. However, a substantial body of work shows that preferences are often non-compensatory: options are rejected when they perform poorly on a critical attribute despite advantages elsewhere \citep{dieckmann2009compensatory, arman2022homogeneous, zhang2025context}. This aligns with bounded rationality accounts that emphasize computational constraints and acceptable trade-offs \citep{simon1990bounded, gershman2015computational}.

Existing approaches capture such behavior through screening rules, threshold-based clipping, and dominance-based restrictions on gain-loss relationships \citep{bierlaire2010analysis, long2014should, dubey2022multinomial, braun2024cone, shukla2020practical}. Unlike threshold-based or lexicographic models, our framework evaluates the balance between gains and losses across attributes, allowing bounded compensation while rejecting excessive sacrifice.

We formulate multi-attribute choice as a trade-off screening process operating on the balance between gains and losses rather than full utility aggregation. Decisions are based on a max-min comparison rule that evaluates whether the largest gain sufficiently offsets the largest loss. The rule is governed by a trade-off tolerance parameter $M$, interpreted as a context-dependent latent quantity controlling acceptable imbalance. The model can be viewed as a probabilistic, pairwise instantiation of bounded trade-off constraints, where \(M\) is an estimable parameter that can vary across contexts, capturing how tolerance to asymmetric trade-offs adapts to the decision environment. Related heuristics such as elimination-by-aspects \citep{tversky1972elimination} and the priority heuristic \citep{brandstatter2006priority} rely on sequential filtering, while dynamic models such as decision field theory \citep{busemeyer1993decision} explain preferences through evidence accumulation. In contrast, our formulation compresses multi-attribute comparisons into extremal gain-loss signals and differs from \cite{braun2024cone, shukla2020practical} by defining a pairwise max-min comparison rule. Through simulation, we characterize behavioral regimes in which the model departs from standard utility-based approaches, including asymmetric trade-off rejection and context-dependent reversals. These patterns define testable hypotheses for future behavioral studies.

\section{Methods}

We model multi-attribute choice as pairwise comparisons. Let \(u, v \in \mathbb{R}^m\) denote two options (higher is better), and define \(d = v - u\). We adopt a bounded trade-off rule, which we refer as $M$-dominance, inspired by dominance-based formulations of bounded trade-offs \citep{shukla2020practical, braun2024cone} with parameter \(M > 0\), under which \(v\) is preferred to \(u\) if
\setlength{\abovedisplayskip}{1pt}
\setlength{\belowdisplayskip}{1pt}
\setlength{\abovedisplayshortskip}{1pt}
\setlength{\belowdisplayshortskip}{1pt}
\begin{equation}
\max_i d_i + M \cdot \min_i d_i > 0.
\end{equation}
Here, $d_i > 0$ denotes a gain and $d_i <0$ a loss for attribute $i$, relative to the reference option $u$. This rule compares the largest gain and largest loss of \(v\) relative to \(u\), accepting options only when gains sufficiently offset the worst deterioration. The parameter \(M\) controls sensitivity to losses relative to gains. Larger values impose stricter screening by requiring gains to more strongly outweigh the worst loss, whereas smaller values allow greater compensation and more permissive trade-offs. To capture variation across decision environments, we consider a context-dependent formulation with \(c \in \mathcal{C}\) and parameter \(M^{(c)}\), yielding $ \max_i d_i + M^{(c)} \cdot \min_i d_i > 0$. 
This treats trade-off tolerance as a context-dependent. The model defines a max-min comparison rule for multi-attribute choice, in which decisions depend on extremal gains and losses rather than full aggregation. The parameter \(M\) (or \(M^{(c)}\)) represents a latent trade-off screening parameter, yielding a simple, computationally efficient rule aligned with bounded rationality and adaptive strategy selection in multi-attribute choice \citep{payne1993adaptive}. The use of max-min can be interpreted as attentional compression \citep{shimojo2003gaze, krajbich2010visual}, where decision makers focus on the most salient gain and loss rather than integrating all attributes. This formulation implies three qualitative patterns: rejection of asymmetric trade-offs even when favored by additive utility, context-dependent preference reversals, and increased sensitivity to extremal attributes. These effects are expected to be most pronounced in mixed trade-off settings involving competing gains and losses. The current formulation assumes that decisions are driven by the most salient gain and loss. Intermediate attributes may still influence choice indirectly through attentional or saliency processes, which we leave to future work.

\section{Results}

\begin{figure}
    \centering
    \includegraphics[width=0.25\textwidth, keepaspectratio]{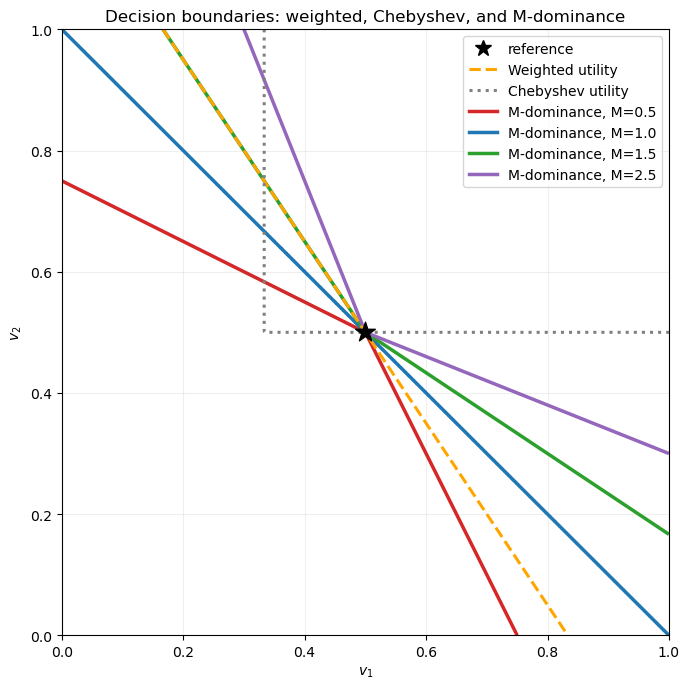}
        \vspace{-4mm}
    \caption{Trade-off boundaries vs \(M\)}
    \vspace{-4mm}
    \label{fig:result_exp1_2}

\end{figure}

\paragraph{Experiment 1: Structural properties of bounded trade-off reasoning}
We compare M-dominance (\(M=1.5\)) with weighted additive and Chebyshev utility on 5000 randomly generated 3-attribute option pairs. M-dominance disagrees with weighted utility on \(86.5\%\) of pairs and with Chebyshev utility on \(74.8\%\), while the two utility models disagree on only \(25.0\%\). It also selects fewer strict preferences (1928 vs.\ 2519 and 2507), rejecting a large fraction of mixed trade-off cases involving simultaneous gains and losses. In such cases, neither option is preferred under the rule. This pattern is consistent with bounded compensatory choice, in which options with strong overall value are nonetheless rejected when they incur a poor outcome on one attribute.  
To further characterize this, we visualize indifference boundaries in a two-attribute space (Fig \ref{fig:result_exp1_2}). Weighted utility produces linear decision boundaries, while Chebyshev induces piecewise boundaries. In contrast, M-dominance generates a family of boundaries that vary systematically with \(M\). Smaller values allow losses to be compensated more easily, while larger values require increasingly large gains to offset losses. This demonstrates that \(M\) controls the geometry of acceptable trade-offs, inducing a continuum of decision regimes.

\paragraph{Experiment 2: Identifiability and context dependence}
We simulate agents under probabilistic bounded trade-off rule with \(M \in \{0.5,1.0,1.5,2.0,3.0\}\) and recover \(M\) via maximum likelihood. The recovered values closely match the ground truth (MAE \(=0.038\), \(r=0.999\)), indicating that trade-off tolerance is reliably identifiable from choice data.We then evaluate context dependence by simulating two environments with distinct  parameters \(M^{(1)}=0.80\) and \(M^{(2)}=2.00\). A context-dependent model accurately recovers both values, while a global model collapses them into a single intermediate estimate. This leads to a substantial improvement in model fit (\(\Delta \log L = 175.65\)) and reveals preference reversals across contexts in \(21.3\%\) of probe comparisons. These results show that trade-off tolerance varies across environments and captures systematic variation in behavior that a single global parameter cannot explain.

\section{Conclusion}

We introduced a bounded trade-off screening model for multi-attribute choice as a computational account of bounded comparison. Decisions are defined by a max--min evaluation of gains and losses, governed by a trade-off tolerance parameter \(M\) that captures sensitivity to asymmetric trade-offs and can vary across contexts. Our simulations show that the model produces preference patterns distinct from utility-based approaches, supports recovery of \(M\) from choice data, and captures context-dependent reversals. These results suggest that multi-attribute choice can be modeled as trade-off screening under bounded tolerance, providing a simple alternative to full utility aggregation. An important next step is to evaluate whether the proposed mechanism explains known behavioral phenomena and provides a better account of human choice than existing compensatory and non-compensatory models.

\section{Acknowledgement}
The authors thank Sabina J. Slomon for her helpful feedback, and several anonymous reviewers for helpful comments. AI-assisted tools was used for language editing and manuscript refinement; all scientific content and claims were developed and verified by the authors. This work was supported by a UKRI Turing AI World Leading Researcher Fellowship on AI for Person-Centred and Teachable Autonomy (grant EP/Z534833/1). 
\printbibliography

\end{document}